\documentclass[b5paper,10pt,abstractoff,DIV=calc,headings=normal,oneside]{scrartcl}
\usepackage[english]{babel}
\usepackage{lmodern}
\usepackage{microtype}
\usepackage{amsmath,amssymb,amsthm,amsfonts,graphicx,etoolbox,scrlayer-scrpage}
\usepackage[noblocks]{authblk}
\usepackage{arydshln}
\usepackage{graphicx}
\usepackage{subcaption}
\usepackage{url}
\usepackage{color}

\makeatletter
\patchcmd{\@maketitle}{\huge}{\Large}{}{}
\patchcmd{\abstract}{\quotation}{}{}{}
\AtBeginEnvironment{abstract}{\noindent\ignorespaces}
\AtEndEnvironment{abstract}{\par\mbox{}}
\newcommand{\shortauthor}{}
\newcommand{\shorttitle}{\@title}
\makeatother
\setkomafont{author}{\small}
\setkomafont{title}{\rmfamily\bfseries}
\setkomafont{disposition}{\rmfamily\bfseries}

\def\AMS#1{\par\noindent \textbf{AMS subject classification: }#1\par}
\newcommand{\acknowledgements}{\par\mbox{}\par\noindent\textbf{Acknowledgements: }}
\newcommand{\keywords}[1]{\par\noindent\textbf{Keywords: }#1}
\rehead[]{\shortauthor}\lohead[]{\shorttitle}\lehead[]{}\rohead[]{}
\ofoot[]{}\ifoot[]{}\recalctypearea
\theoremstyle{plain}

\theoremstyle{definition}

\theoremstyle{remark}

\renewenvironment{abstract}{\bigskip\noindent\begin{minipage}{\textwidth}\setlength{\parindent}{15pt}\paragraph{Abstract:}}{\end{minipage}}

\begin{document}


\renewcommand{\shortauthor}{V. Gallego}
\renewcommand{\shorttitle}{Bradley-Terry Models in Text-To-Image Classification and Generation}

\title{Fast Adaptation with Bradley-Terry Preference Models in Text-To-Image Classification and Generation}

\author[1]{Víctor Gallego\thanks{Corresponding author: victor.gallego@komorebi.ai}}
\affil[1]{Komorebi AI Technologies}

\maketitle

\begin{abstract}

Recently, large multimodal models, such as CLIP \cite{pmlr-v139-radford21a} and Stable Diffusion \cite{rombach2022highresolution} have experimented tremendous successes in both foundations and applications. However, as these models increase in parameter size and computational requirements, it becomes more challenging for users to personalize them for specific tasks or preferences.
In this work, we address the problem of adapting the previous models towards sets of particular human preferences, aligning the retrieved or generated images with the preferences of the user. We leverage the Bradley-Terry preference model \cite{bradley1952rank} to develop a fast adaptation method that efficiently fine-tunes the original model, with few examples and with minimal computing resources.
Extensive evidence of the capabilities of this framework is provided through experiments in different domains related to multimodal text and image understanding, including preference prediction as a reward model, and generation tasks.

\end{abstract}

\smallskip
\AMS{62H30, 62H35} 


\section{Background}
\paragraph{Contrastive models.}Given a textual prompt, such as a sentence in English $t$, and an image $I$, both data points can be processed by a contrastive model such as CLIP \cite{pmlr-v139-radford21a}  to compute a latent representation of the inputs. That is, the text $t$ is feed through an encoder model (typically a transformer-based neural model) arriving at a text embedding $x = f_{\theta_1}(t)$, and the same for the image $I$, arriving at an image embedding $y = f_{\theta_2}(I)$. Both embeddings lie on a shared high-dimensional space, $x, y \in \mathbb{R}^d$, with $d=768$ or $1024$, depending on model size. In this shared space, we can compute the similarity $s(x,y)$ between the text and the image, using the dot product (assume the embeddings are $\ell_2$-normalized, and $x, y$ are column vectors):
\begin{eqnarray}\label{contrastive}
s(x, y) = x^\intercal y.
\end{eqnarray}
Using this metric, we can do several tasks, such as retrieving the closest image to a given textual prompt, or classifying images based on a set of texts, etc.

Large-scale models such as CLIP have been massively pretrained over vast amounts of (text, image) pairs crawled from the internet. As users interact with them, it is of great interest to develop techniques to further fine-tune these models, personalizing them towards their user preferences, using as little data as possible. Thus, there is growing interest in creating datasets of human preferences, such as SAC \cite{pressmancrowson2022} or DiffusionDB \cite{wangDiffusionDBLargescalePrompt2022}, and models further adapted to human preference, such as PickScore \cite{kirstain2023pick}.

\paragraph{Bradley-Terry models.} Assume a pair of items $i, j$ sampled from a population, the Bradley-Terry (BT) model \cite{bradley1952rank} estimates the probability that the pairwise comparison $i \succ j$, which indicates a strong preference of $i$ over $j$, is true as:
\begin{eqnarray}
    P(i \succ j) = \dfrac{\exp s_i}{\exp s_i + \exp s_j},
\end{eqnarray}
with $s_i$ being a latent variable representing the strength of item $i$.

\section{The CLIP-BT framework}

We now introduce our main contribution: using the BT model to fine-tune a contrastive model like CLIP in an efficient way.
Given a textual embedding $x$ and two image embeddings $y_1$ and $y_2$, the probability $\hat{p}_1$ of choosing image $y_1$ versus $y_2$ can be computed using the Bradley-Terry model with the similarity scores as the strength variables,
$
\hat{p}_1 = \frac{\exp s(x, y_1)}{\exp s(x, y_1) + \exp s(x, y_2)}.
$
Assuming that we do not allow ties; that is, for any pair of images, we can sort the indices to have $y_1 \succ y_2$, we can define the loss function as the negative log-likelihood:
\begin{eqnarray}\label{preference_loss}
\mathcal{L}_{pref}(x, y_1, y_2) = - \log \hat{p}_1.
\end{eqnarray}
The case of having $N$ pairs of preference data, $\lbrace y_{i1} \succ y_{i2} \rbrace_{i=1}^N$ can be straight-forwardly extended by simply averaging the previous loss over each pair. 

Using standard automatic differentiation libraries, we can compute the  gradients of the previous loss with respect to the encoder parameters, $\theta_1, \theta_2$, to optimize the model towards the preferences of the user. However, for many applications we can instead take the gradient with respect to the textual prompt $x$, involving much less computational burden. Thanks to the linearity of (\ref{contrastive}), after a few simplifications we arrive at
\begin{eqnarray}\label{adapt}
\nabla_x \mathcal{L}_{pref}(x, y_1, y_2) = (\hat{p}_1 - 1)(y_1 - y_2).
\end{eqnarray}
That is, given a textual query $x$, and a pair of user preferences $y_1 \succ y_2$, we compute a refined prompt $x'$ with just one iteration of gradient descent, i.e. $x' = x - \epsilon\nabla_x \mathcal{L}_{pref}(x, y_1, y_2)$, with $\epsilon > 0$ being a learning rate. Intuitively, this update rule steers the embedding $x$ towards the preferred image embedding $y_1$, and far from the negative one, $y_2$. 

Note that with this approach, we do not need to modify the underlying encoder models, thus this is a fast preference adaptation method, that just requires modifying the embedding $x \in \mathbb{R}^d$. Also notice that the previous gradient only involves subtracting the two image embeddings, and computing the probability $\hat{p}_1$, so it can be computed efficiently, compared to fine-tuning the parameters of the encoder models. These are the main benefits of our approach.
\paragraph{What to do when we do not have pairwise preference data?} If we do not have a set $\lbrace y_{i1} \succ y_{i2} \rbrace$ of both positive and negative samples, but rather just a set of highly-preferred images $\lbrace y_{i1} \rbrace$, we cannot use the previous BT loss. But we can still adapt the embedding $x$ towards that set, by maximizing the similarity (\ref{contrastive}) with one iteration of gradient descent: $x' = x + \epsilon \frac{1}{N} \sum y_{i1} $. This is a baseline that we will use through the experiments.

\section{Experiments}


\subsection{Classification task as a preference model}
DifussionDB is a recent large-scale dataset consisting in triplets $(x, y_1, y_2)$, in which users voted whether $y_1 \succ y_2$ or the opposite. We randomly split 2000 samples as an evaluation set, and experiment with different training sets of sizes from 1 to 50 preference pairs. As the textual query, we take the embedding $x$ corresponding to the text $t=\texttt{An amazing and high-quality image}$, with the objective of predicting, for each pair, the preferred image in terms of quality. We experiment with CLIP models of different sizes: CLIP-L (430M parameters), CLIP-H (1B parameters), and CLIP-G (2.5B parameters).

\begin{figure}[!ht]
\begin{minipage}{0.4\textwidth}
  \centering
\resizebox{\columnwidth}{!}{%
  \begin{tabular}{rc}
  \hline
  Config.               & Accuracy \\ \hline
  CLIP-L-original & $0.548$ \\
  CLIP-L-positive & $0.695 \pm 0.003$ \\
  CLIP-L-BT & $\mathbf{0.754 \pm 0.007}$ \\ \hdashline
  CLIP-H-original           & $0.644$ \\
  CLIP-H-positive        & $0.742 \pm 0.007$ \\
  CLIP-H-BT         & $\mathbf{0.788 \pm 0.003}$ \\ \hdashline
  CLIP-G-original          & $0.624$ \\
   CLIP-G-positive           & $0.731 \pm 0.007$ \\
  CLIP-G-BT         & $\mathbf{0.772 \pm 0.004}$ \\ \hline
  PickScore \cite{kirstain2023pick}          & $0.738$  
  \end{tabular}
  }
\end{minipage}%
\hspace{0.7cm} 
\begin{minipage}{0.5\textwidth}
  \centering
  \includegraphics[width=6cm]{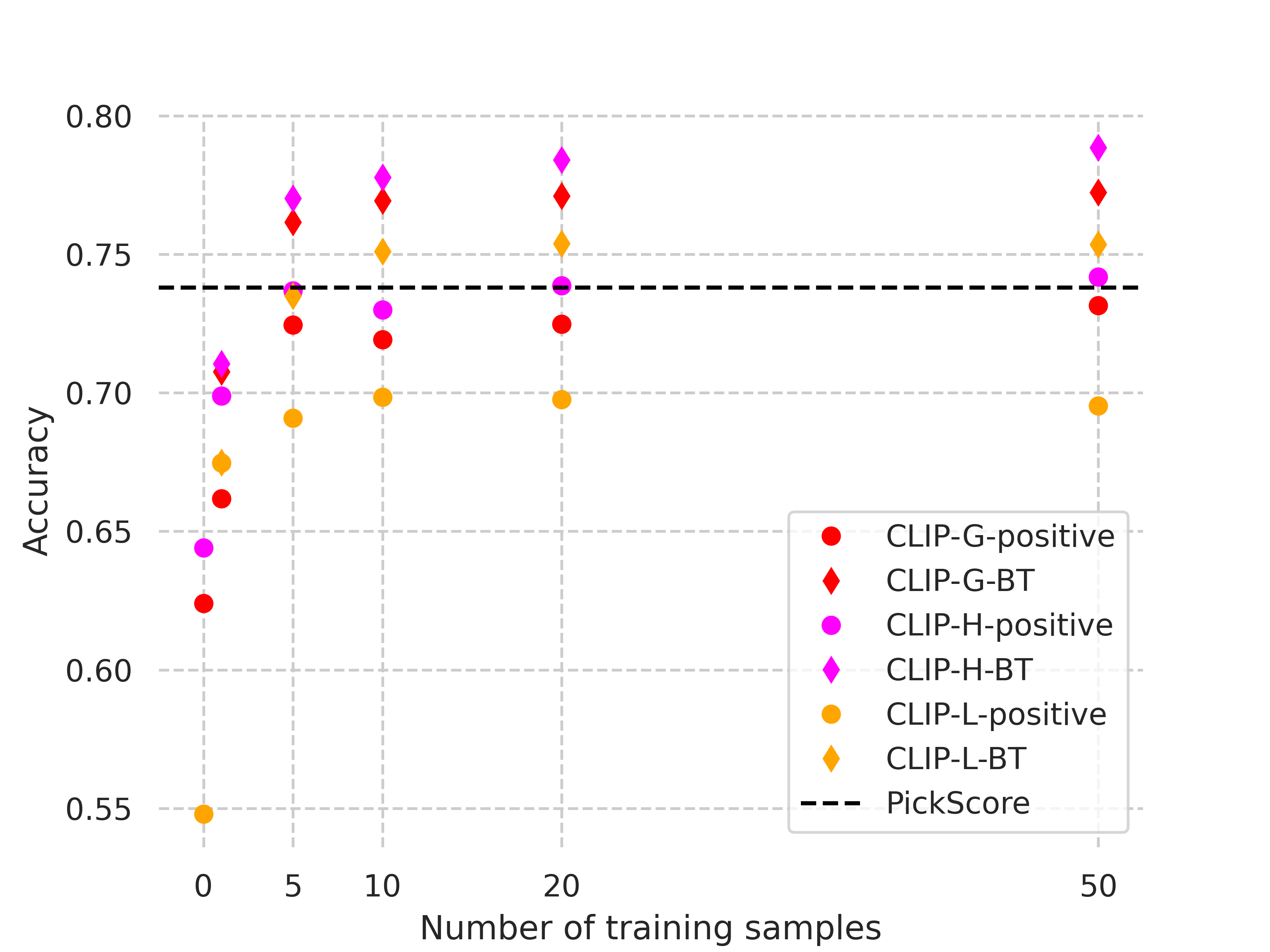}
\end{minipage}
\caption{Preference classification results}\label{fig:classification}
\end{figure}

Results for training set of size $N = 50$ are shown in Figure \ref{fig:classification} left. For each pair of testing images, we compute the preferred image by the model; and as the accuracy metric, we consider a match if that predicted preferred image matches the preferred one in the ground truth. 
Accuracy std's are computed over 10 different random training sets. For each model size, we report accuracies with the original model (original), a fast adaptation based on just using the positive samples of each training pair (positive), and our fast adaptation gradient from (\ref{adapt}) (BT). We also include the results of a recent state-of-the-art model, PickScore \cite{kirstain2023pick}, which was fine-tuned on a different preference dataset. Next, in Fig. \ref{fig:classification} right we show the mean accuracies while varying the size of the training set from 0 (no adaptation) to 50.

The previous results suggest that just by using a small set of preference data, we can achieve superior results that the fully fine-tuned model PickScore, which is a CLIP model trained over a similar dataset of highly-scored images. Also, the BT loss clearly improves on our baseline that only uses positive samples, which is expected as we are giving it more learning signal thanks to the negative samples.
\subsection{Text-to-image generation}
Stable Diffusion \cite{rombach2022highresolution} is a state-of-the-art model for text-to-image generation. Internally, it uses the CLIP model to compute a latent representation from a given text prompt, which is later used by a diffusion process to generate an image $y$. We are interested in performing an adaptation of the model towards specific aesthetics preferred by the user, specified as a set of image preferences. To obtain such a set of image preferences, the SAC dataset contains images with human rankings \cite{pressmancrowson2022}. A recent work of us already tackled the problem of steering the model towards a set of images, but without negative preferences, just using images with high scores \cite{gallego2022personalizing}. Here, we also use images with low scores, and form a preference dataset by sampling pairs from the previous two groups, $\lbrace y_{i1} \succ y_{i2} \rbrace_{i=1}^N$, with $\lbrace y_{i1} \rbrace $ being the set of highly-scored images, and $\lbrace y_{i2} \rbrace $ the other one, with the lowest scores. We also use the loss function from (\ref{preference_loss}), and adapt the CLIP model for between $T=5$ and $T=7$ optimization steps of gradient optimization. The benefit is that we do not need to retrain the full generative model, as the optimization is done at inference time, only adding a small computing time overhead.

\begin{figure}[!ht]
\begin{minipage}{0.3\textwidth}
  \centering
  \resizebox{\columnwidth}{!}{%
  \begin{tabular}{rcc}
\hline
Config & Win rate   \\ \hline
CLIP-L-original     & 0.267    \\
CLIP-L-positive    & 0.273  \\
CLIP-L-BT      & \textbf{0.460}
\end{tabular}
}
\end{minipage}%
\hspace{0.7cm} 
\begin{minipage}{0.69\textwidth}
\centering
    \setlength{\abovecaptionskip}{1.5pt}
    \setlength{\belowcaptionskip}{-3.5pt}
    \setlength{\tabcolsep}{0.55pt}
    \renewcommand{\arraystretch}{1.0}
    {
    
    \begin{tabular}{c}
    
    \begin{tabular}{c c @{\hskip 30pt}  c c c }
     \\

    \multicolumn{2}{c}{\raisebox{0.01\linewidth}{\begin{tabular}{c@{}c@{}} Textual prompt \end{tabular}} }   &
    \multicolumn{3}{c}{\raisebox{0.01\linewidth}{\begin{tabular}{c@{}c@{}c@{}} Generated samples \end{tabular}} }  \\ \hline \\[-1pt]

    \multicolumn{2}{l}{\raisebox{0.045\linewidth}{\tiny\begin{tabular}{l@{}l@{}}  A pirate ship, sepia coloring,\\ hyper-detailed, dusk,\\ 4k octane render \end{tabular}} }   &
    \includegraphics[width=0.12\linewidth,height=0.12\linewidth]{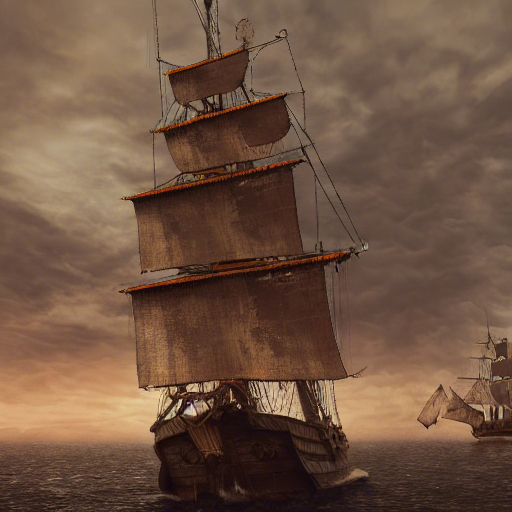} & 
    \includegraphics[width=0.12\linewidth,height=0.12\linewidth]{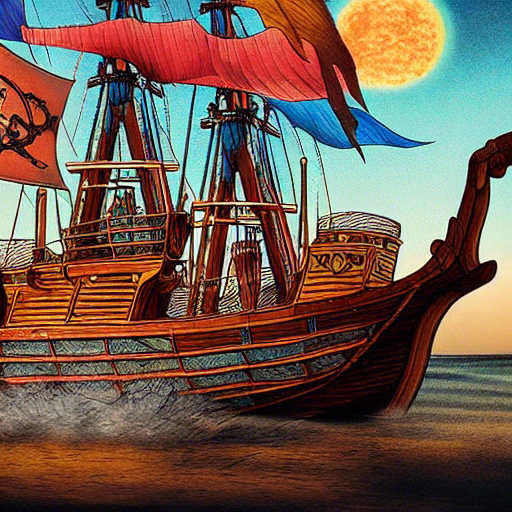}& 
    \includegraphics[width=0.12\linewidth,height=0.12\linewidth]{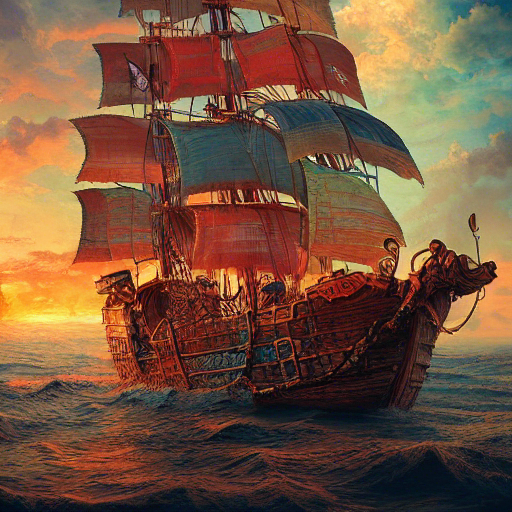} \\[2pt]
    
    \multicolumn{2}{l}{\raisebox{0.045\linewidth}{\tiny\begin{tabular}{l@{}l@{}}  A still life of flowers,\\ volumetric lighting
 \end{tabular}} } &
    \includegraphics[width=0.12\linewidth,height=0.12\linewidth]{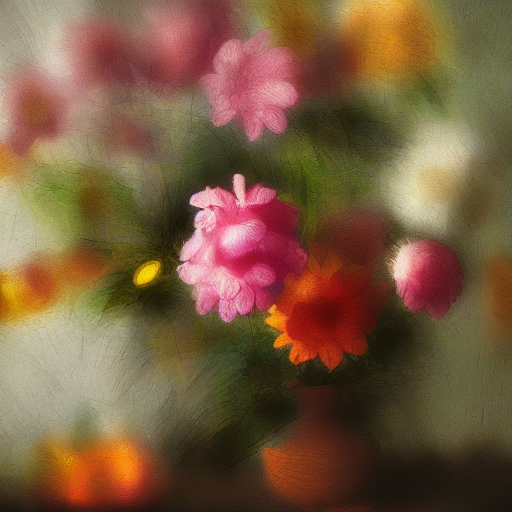} & 
    \includegraphics[width=0.12\linewidth,height=0.12\linewidth]{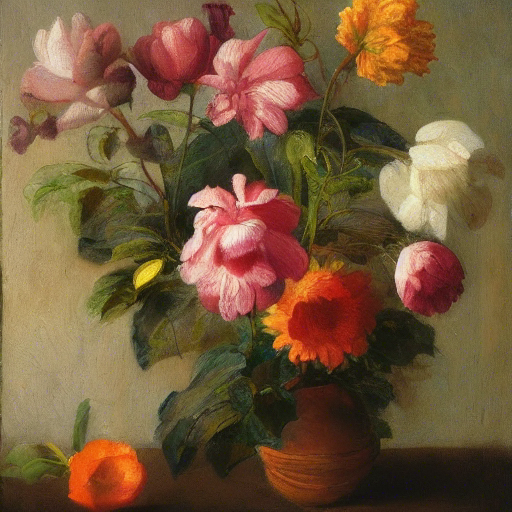} &
    \includegraphics[width=0.12\linewidth,height=0.12\linewidth]{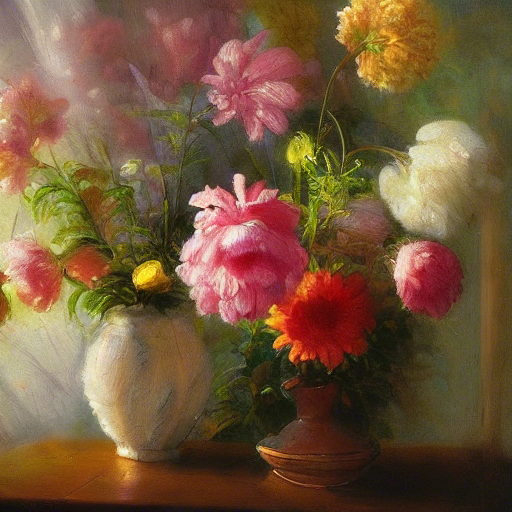} \\[2pt] \hline \\[-2pt]

    \end{tabular}
    
    \end{tabular}

    }
\end{minipage}
\caption{Results for the text-to-image generation experiment. Left: human evaluation results. Right: Stable Diffusion generations. For each prompt, each generation is sampled using CLIP-L-original, CLIP-L-positive, and CLIP-L-BT, respectively.}\label{fig:generation}
\end{figure}

For a fixed set of 20 textual prompts, we generate several image samples, with the following three variants to the underlying CLIP model:
CLIP-L-original, CLIP-L-positive (adapted only with the positive set of images \cite{gallego2022personalizing}); and CLIP-L-BT (adapted using both positive and negative samples, and loss from (\ref{preference_loss})).
We evaluate the quality of each variant with a human evaluation using two annotators. For each prompt, we generate three images using the three previous variants, and each annotator voted their most preferred generation. Results are shown in Fig. \ref{fig:generation} left (Win rate), confirming the superiority of the optimization via the BT loss. Lastly, Figure \ref{fig:generation} right shows a random pair of samples for each variant.

\section{Conclusions and further work}
We have proposed a fast adaptation, gradient-based method, which leverages the linearity in the CLIP scoring function and the BT model. Since we do not need to store modified weights for the CLIP model, the technique introduced here is scalable to regimes of many different preference profiles.

The applicability of the technique was demonstrated with experiments in preference classification and improving text-to-image generation towards user preferences. An interesting line of work would be to adapt more general preference models, such as the Plackett-Luce variant, to account for more than two preference levels, e.g., $y_1 \succ y_2 \succ y_3$.

\bigskip
\acknowledgements{The author acknowledges support of the Torres-Quevedo postdoctoral grant PTQ2021-011758.}


\bibliographystyle{abbrv}
\bibliography{bib}

\end{document}